# Exploring Runtime Decision Support for Trauma Resuscitation

Keyi Li[1], Sen Yang[2], Travis M. Sullivan[3], Randall S. Burd[3], Ivan Marsic[1]

[1]Electrical and Computer Engineering Department, Rutgers University, [2]LinkedIn, [3]Children's National Hospital

{kl734, sy358, marsic}@rutgers.edu, {tsullivan, rburd}@childrensnational.org

## ABSTRACT

AI-based recommender systems have been successfully applied in many domains (e.g., e-commerce, feeds ranking). Medical experts believe that incorporating such methods into a clinical decision support system may help reduce medical team errors and improve patient outcomes during treatment processes (e.g., trauma resuscitation, surgical processes). Limited research, however, has been done to develop automatic data-driven treatment decision support. We explored the feasibility of building a treatment recommender system to provide runtime next-minute activity predictions. The system uses patient context (e.g., demographics and vital signs) and process context (e.g., activities) to continuously predict activities that will be performed in the next minute. We evaluated our system on a prerecorded dataset of trauma resuscitation and conducted an ablation study on different model variants. The best model achieved an average F1-score of 0.67 for 61 activity types. We include medical team feedback and discuss the future work.

## 1 Introduction

Critically injured patients have a four-fold higher risk of death from medical errors compared to non-injured patients [11]. During the initial evaluation and treatment of an injured patient, a multidisciplinary medical team rapidly identifies and manages life-threatening injuries. Despite a universal protocol to rapidly evaluate and treat injured adults and children, avoidable errors are common, and deviations from this protocol have been observed in up to 85% of trauma resuscitations [1, 2, 12]. Our research explores the feasibility of a data-driven decision support system to improve the medical process performance. Our system provides next-minute treatment predictions based on a deep neural network and data-mining techniques (Figure 1).

Our problem differs from traditional sequential recommender problems [5] in two ways. First, both patient and process contextual information play essential roles in predicting the activities to perform in the next minute. For example, if a patient arrived with a blunt injury, the medical team may perform activities to immobilize the cervical spine while simultaneously performing blood transfusion-related activities if the patient is also hypotensive. Process context (i.e., the past treatment activities) is also crucial in predicting the future activities. Both long-range and short-range dependencies between activities are important in predicting the activities to perform next.

Second, our runtime system in parallel gives binary prediction for all possible activity labels. During trauma resuscitations at urban trauma centers, multidisciplinary team members perform most treatment activities in parallel instead of sequentially. Unlike traditional sequential recommender systems, our system does not rank-order and output the top $k$ most likely activities to perform, but rather every minute during the process outputs a set of activities for the medical team.

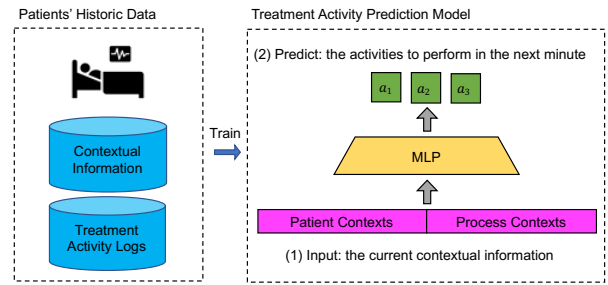

**Figure 1: Overview of our context-aware runtime decision support system development.**

We designed a multi-label predictive network to handle these differences. The network needed to incorporate contextual information and continually output multiple predictions. Our dataset consisted of activity labels over time intervals during the entire resuscitation. We sampled each resuscitation case at regular one-minute steps and labeled each sample with the activities that occurred during the next minute. The goal was to predict these activities based on the contextual features associated with each sample. We used a conceptually simple and efficient model architecture built entirely on multi-layer perceptrons (MLPs) to do the multi-label classification.

The main contributions of our work are:

- A case study to explore a prediction-based decision support for the trauma resuscitation process.
- An MLP-based network that aggregates both patient and process contextual information as input features and outputs multi-label predictions.

## 2 Method

**Features.** The feature set includes patient context features and process context features. In the resuscitation processes, patient contexts include static context (i.e., patient's demographics and their vital signs on arrival) and dynamic context (i.e., the continuous vital sign measurement). Static context is recorded initially at the patient arrival (represented by vector $s = [s_1, ..., s_h]$, where each dimension represents one attribute). The dynamic context is recorded manually from time to time during the process by checking the vital monitors ($d^t = [d_1, ..., d_m]$, where each dimension represents one attribute, and $t$ is the recorded timestamp).

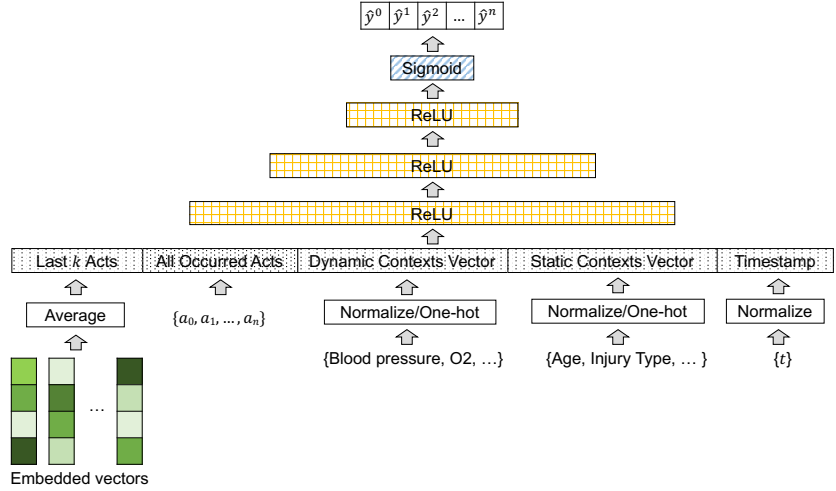

**Figure 2: The architecture of our prediction model for next treatment activities. All contextual features are encoded into a 1D vector and concatenated (the dotted bar). After the fully connected ReLU hidden layers, the output features go through the Sigmoid function and return the predictions for each activity label.**

Both static and static context contains categorical features (e.g., injury type) and numerical features (e.g., heart rate). We encoded the categorical features using one-hot vectors. The numerical features are normalized between [0,1] using linear scaling $x' = \left(\frac{x-x_{min}}{x_{max}-x_{min}}\right)$, where $x_{max}$ and $x_{min}$ are the maximal and minimal value of that feature in the dataset. The one-hot vectors and numerical vectors are then concatenated to represent the dynamic contexts ($X_d \in \mathbb{R}^M$) and static contexts ($X_s \in \mathbb{R}^H$), where $M$ is the number of categorical and numerical data types for dynamic context, and $H$ is the corresponding number of data types for static context. Because dynamic context data are updated at random times, we continued using the most recent dynamic context until the next update.

Each resuscitation case is recorded as an activity log, which provides the process contextual information. The logs contain activity traces indicating the start and end times for each activity. To extract the short- and long-range dependencies between the activities, we selected the most recent $k$ activities (we chose the last five in this work) to capture the short-range dependencies ($X_{sa}$), and aggregated all past activities to capture the long-range dependencies ($X_{la}$). The last five activities were usually performed in the previous two to three minutes in our dataset. If more than five activities were recorded in one minute, we randomly selected five of these activities. We embedded these five activities into dense tensors and applied average pooling ($X_{sa} \in \mathbb{R}^k$) [7, 10]. All past activities from time 0 to time $t$ are collected in $X_{la}$, which is a 1D vector of length $n$, where $n$ is the number of activity types, and each element is a binary indicator of whether the activity has occurred ($X_{la} \in \mathbb{R}^n$). Considering that the medical team members usually perform some activities at specific point in the process (e.g., vital sign are often obtained at the beginning of a trauma resuscitation), the timestamp $t$ is also important for activity prediction. Timestamp $t$ is calibrated based on patient arrival time and is used as a numerical feature ($X_t \in \mathbb{R}^1$).

**Model Architecture.** The activity predictions for time $t$ are made based on the patient context (static and dynamic) at time $t$, and the process context up to time $t$. We concatenated all the context features into $X$ and feed it to the network. The architecture of our MLP-based model is shown in Figure 2. The concatenated feature vector $X$ is fed into the fully connected hidden layers of Rectified Linear Units (ReLU) [3]. We used the batch normalization method to improve the model convergence and prediction accuracy [6]. Given an input sample $X$ at time $t$, the ground-truth label $y$ of $X$ contains the activities being performed during the next minute, denoted as $y = [y^1, y^2, ... y^n]^T$, where each binary element $y^i$ indicates whether the activity $a_i$ is performed ($y^i = 1$) or not performed ($y^i = 0$). Thus, the last layer of our model projects the output of the ReLU layers into $n$ labels, which is followed with a Sigmoid function. Our model is represented as:
$$\hat{y} = f_{mlp}(X; \theta_{mlp})$$
where $\hat{y} = [\hat{y}^1, \hat{y}^2, ... \hat{y}^n]^T \in \mathbb{R}^n$ are the predicted probabilities for each activity label, and $\theta_{mlp}$ are the model parameters.

**Label Imbalance.** The activity labels are usually highly imbalanced in the trauma resuscitation dataset. While some general treatment activities occur in nearly every log, other activities only occur based on the severity of the patient's injury. When the logs are sampled at one-minute points, the long-duration activities dominate the sampled data. The high frequency of these labels might dominate the training process and impact the model performance for infrequent or short activities.

**Table 1: The prediction performances (weighted F1-score and samples F1-score) for MLP-based model (with different context combinations).**

| Contexts | | Weighted F1-score | Samples F1-score |
|---|---|---|---|
| Last $k$ activities | | 0.625 | 0.491 |
| All occurred activities | ←Process context | 0.577 | 0.385 |
| Last $k$ acts + All occurred activities | | 0.654 | 0.444 |
| Dynamic patient context | | 0.427 | 0.211 |
| Static patient context | ←Patient context | 0.250 | 0.110 |
| Dynamic patient context + Static patient context | | 0.448 | 0.212 |
| Timestamp | | 0.362 | 0.189 |
| Last $k$ acts + All occurred acts + Dynamic patient context | | 0.664 | 0.450 |
| Last $k$ acts + All occurred acts + Static patient context | | 0.662 | 0.523 |
| Last $k$ acts + All occurred acts + Dynamic context + Static context | | 0.665 | 0.551 |
| Last $k$ acts + All occurred acts + Timestamp | | 0.655 | 0.454 |
| Last $k$ acts + All occurred acts + Dynamic context + Static context + Timestamp | | 0.671 | 0.556 |

In the model training phase, we applied focal loss [8] to help with the label imbalance. To adapt focal loss in multi-label classification problem, in each training batch, we have:

$$FL = \frac{\sum_{j=1}^{b} \sum_{i=1}^{n} -\alpha^i (1 - \hat{y}_j^i)^\gamma \log(\hat{y}_j^i)}{b}$$

where $b$ is the batch size, $\alpha^i$ and $\gamma$ are the hyper-parameters. Focal loss assigns larger losses to the "hard-to-predict" labels. Larger losses force the model to focus and learn more from the bad predictions. For the hyper-parameters, we set $\alpha^i$ as the label weight (i.e., $1 - f_{a_i}$, where $f_{a_i}$ is the fraction of label $a_i$ in the dataset), and we set $\gamma$ to 2.

**Evaluation Metrics.** Our system should predict medical activities correctly and at the right time. Given a large set of activities, of which only a few will be performed at any time during the process, our dataset is highly imbalanced with many true negatives compared to true positives. Therefore, we used the weighted F1-score to measure the predictive performance of our model. To measure the multi-label classification performance, we analyzed the samples F1-score.

**Inference.** In the model inference phase, the output $\hat{y}$ of the Sigmoid layer is further classified into binary predictions for each activity label. We assigned each activity label an individual probability threshold to decide if the activity would occur in the next minute. We applied the threshold-moving strategy to identify the optimal threshold [4], i.e., identifying the best threshold for each activity label by maximizing the F1-score in the precision-recall curve.

## 3 Experiments

**Dataset and Experiment Set Up.** The use of medical data for this study was approved by the Institutional Review Board at our hospital. Medical experts manually coded 201 trauma resuscitation cases into activity logs based on surveillance videos. The total duration of the 201 cases is 5708 minutes, or 28±13.93 minutes per case. We have 2024 dynamic-context vectors recorded in the 201 cases, or 10±6.78 vectors per case. Each case had one static context vector. There were 61 types of activities recorded in the activity logs. Each dynamic-context vector contained six types of features (numerical features: heart rate, respiratory rate, systolic blood pressure, diastolic blood pressure, oxygen saturation; and one categorical feature: FiO2), and each static vector contained six types of features (numerical features: age, Glasgow Coma Scale (GCS) score, Abbreviated Injury Scale (AIS) score, heart rate, systolic blood pressure; and one categorical feature: injury type).

We divided the 201 cases into training, validation, and test sets in a 161:20:20 ratio. Each set was preprocessed separately, and the total number of data points by minute was 5708. Each sample contained the contextual features and was labeled with the ongoing activities. If a sample was missing some elements of the context vector, we padded them with dummy tokens.

We embedded each of the 61 types of activities into a 16-element vector. The concatenated input was fed into the ReLU layers. A final linear layer with the Sigmoid function projected the output into 61 dimensions to reflect the predictions. We set the training batch size to 64 and a learning rate at 0.0001. Early stopping was applied when the validation loss converged.

**Baselines.** Our context-aware prediction model used five different types of contextual information for predicting the next activities (the dotted bar in Figure 2). We set the baselines by combining different contexts to train the model. We evaluated the importance of each context type for activity prediction by comparing the performance of different context combinations.

**Results.** The prediction performances for different context combinations are shown in Table 1. For both weighted and samples F1-scores, our model using all context features outperformed other models (the full MLP model highlighted in Table 1). We found that the process context (i.e., last $k$ activities, all occurred activities) better predicted the next activities than the patient contexts (dynamic and static). It appears that the process-based contextual information (the process workflow) can best predict the next activities, and the patient context contributes only relatively small prediction improvement. Because medical resuscitations are performed by a multidisciplinary team who perform tasks simultaneously, medical providers may prefer to stick with their routine workflow.

We also compared the model trained on patient vs. process contexts individually. We found that the last $k$ activities outperformed all past activities and timestamps. Also, the dynamic patient context outperformed the static one. These observations

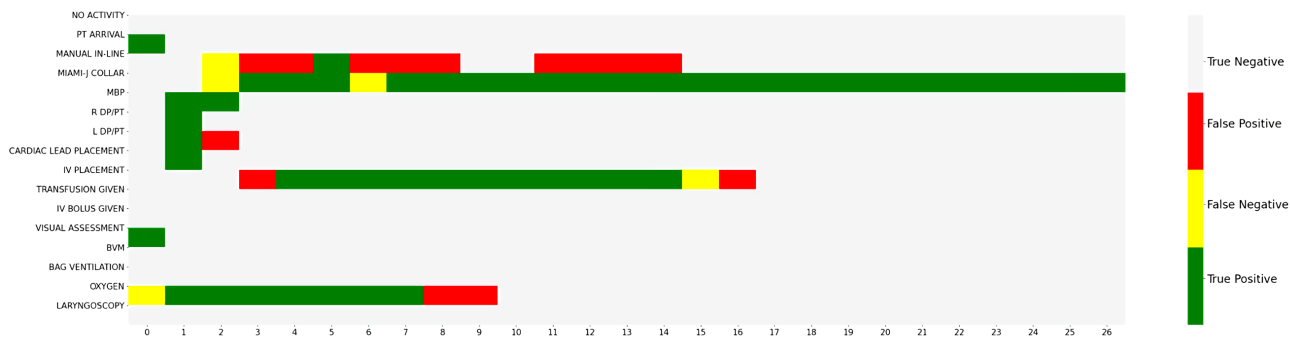

**Figure 3: The prediction performance in one full resuscitation case using the MLP model with full context information. The x-axis is the timestamp (in minutes), and the y-axis shows the activities for which F1-score > 0.5. (Better view in color)**

showed that the current context (last $k$ activities and dynamic patient data) contributed more than the long-range context (all past activities and static patient data), demonstrating the importance of recent/dynamic observations for predicting the next steps.

Figure 3 illustrates one of our test cases using the MLP model trained with all contexts. The horizontal bars show the correctness of the predicted next activities—shown only for the activities for which F1-score on the test dataset was greater than 0.5. We observed more true positive predictions at the beginning of the process. For example, from 0 to 2 minutes, activities like pulse check (R DP/PT, and R DP/PT), blood pressure check (MBP), and visual assessment were correctly predicted. In our dataset, these activities were more likely to be observed at the start of a resuscitation because the medical teams usually perform these general assessments right after a patient's arrival.

We also observed many true positive predictions for activities with long durations, such as Miami-J collar and oxygen, or needed to be repeated in different locations or because of failed attempts, such as intravenous access (IV placement). These activities were easier to predict because the next-minute performance usually depended on the previous minute. For medical processes performed by multidisciplinary teams, the continuous predictions for activities that need to be repeated, like IV placement, might help with reminders for the team members.

## 4   Discussion

Our experiment results showed that our prediction model achieved improvement using comprehensive contextual information. The current prediction accuracy is unsatisfactory because, among the 61 activity types in our dataset, only 16 achieved an F1-score higher than 0.5. We found that the first occurrence was not predicted for some long activities with a good F1-score (e.g., oxygen), but accurately predicting the first occurrence would be more helpful than predicting the persistence of an activity. However, increasing the prediction accuracy should not be the only goal. Given that avoidable errors and deviations are commonly observed in trauma resuscitations [1, 2, 12], it is likely that our dataset contains such errors. A high-performing system trained on such data would keep recommending erroneous steps. On the other hand, obtaining the ground truth about medical errors is laborious and requires medical expertise. We believe that comparing the predictions made by our system trained on different contexts can help pinpoint the potential medical errors that medical experts quickly check. For example, in many resuscitation cases, we observed that patients' dynamic vital signs changed significantly after some essential activities (e.g., blood pressure increased after blood transfusion). The vital signs in these cases did not help the prediction prior to activity occurrence. We believe further analysis of the difference between the predictions based on process context vs. patient context can reveal whether providers may have followed their routine workflow and paid insufficient attention to the patient context.

Although the system will output new set of predictions every minute, most of the predicted activities will be repeated from the previous interval, so the observer will not notice the difference. On the other hand, our medical experts expressed opinion that the 1-minute should be used because for some critical and timely interventions like blood transfusion and intubation, a timely prediction is important. Different prediction and display intervals may be needed for different activity types for an effective run-time decision support.

Some potential future directions include: (1) collecting more data, (2) leveraging process mining techniques [9] to predefine a standard trauma resuscitation workflow and reduce the recommendation search space, (3) focusing on predicting a subset of critical activities, instead of all activities, and (4) analyzing the differences between the predictions based on process context vs. patient context for potential medical errors.

## ACKNOWLEDGEMENT

This work is supported by the U.S. National Institutes of Health/National Library of Medicine under grant number R01LM011834.